\documentclass[12pt,letterpaper]{article}
\usepackage{setspace}
\usepackage[utf8]{inputenc}
\usepackage[T1]{fontenc}
\usepackage{graphicx}
\usepackage{amsmath}
\usepackage{amsfonts}
\usepackage{amssymb}
\usepackage{amstext}
\usepackage{makecell}

\usepackage[hmargin=0.8in,vmargin={1.2in,1.1in}]{geometry}
\usepackage{kpfonts}
\usepackage{rotating}
\usepackage{booktabs}
\usepackage{csquotes}
\usepackage{amsbsy}
\usepackage{eqnarray}
\usepackage{lipsum}
\usepackage{dsfont}
\usepackage[affil-it]{authblk}
\usepackage{float}
\usepackage{tabularx}
\usepackage{multirow}
\usepackage{multicol}
\usepackage{color}
\usepackage{url}
\usepackage{soul}
\usepackage[official]{eurosym}
\usepackage[colorlinks,hypertexnames=false]{hyperref}
\usepackage[dvipsnames,table,xcdraw,svgnames]{xcolor}
\usepackage{subfig}
\usepackage{longtable}
\usepackage{verbatim} 
\usepackage{fancyvrb} 
\usepackage{longtable,booktabs} 
\usepackage{adjustbox}          
\usepackage{array} 
\usepackage{hyperref}
\newcolumntype{C}[1]{>{\centering\arraybackslash}m{#1}}

\definecolor{deeppink}{rgb}{1.0, 0.08, 0.58}

\usepackage[font=footnotesize,labelfont=large, labelsep=period]{caption} 

\graphicspath{{Figures/}}

\makeatletter
\renewcommand\footnotesize{%
	\@setfontsize\footnotesize\@ixpt{11}%
	\abovedisplayskip 8\p@ \@plus2\p@ \@minus4\p@
	\abovedisplayshortskip \z@ \@plus\p@
	\belowdisplayshortskip 4\p@ \@plus2\p@ \@minus2\p@
	\def\@listi{\leftmargin\leftmargini
		\topsep 4\p@ \@plus2\p@ \@minus2\p@
		\parsep 2\p@ \@plus\p@ \@minus\p@
		\itemsep \parsep}%
	\belowdisplayskip \abovedisplayskip
}
\makeatother

\usepackage[sortcites = false, 
style=authoryear-icomp, 
backend = biber, 
sorting=nyt,
hyperref=true,
maxcitenames=1]{biblatex} 
\renewbibmacro{in:}{}
\DeclareLanguageMapping{american}{american-apa}

\usepackage{xcolor}
\definecolor{bloodorange}{HTML}{D15500}
\definecolor{bloodorange2}{HTML}{F15000}
\definecolor{darkgreen}{HTML}{004d00}


\AtBeginDocument{%
\hypersetup{
  citecolor=darkgreen,    
  linkcolor=bloodorange,  
  urlcolor=MidnightBlue   
}
}

\usepackage{etoolbox}
\newbool{inboldface}
\setbool{inboldface}{false}
\let\oldtextbf\textbf
\renewcommand{\textbf}[1]{%
  \setbool{inboldface}{true}%
  \oldtextbf{#1}%
  \setbool{inboldface}{false}%
}
\let\oldtextsc\textsc
\renewcommand{\textsc}[1]{%
  \ifbool{inboldface}%
    {\oldtextsc{\textcolor{darkgreen}{#1}}}%
    {\oldtextsc{#1}}%
}

\title{
    \vspace{-2cm} 
    {\fontsize{20pt}{25pt}\selectfont \textbf{{Beyond Public Access in LLM Pre-Training Data}}}\\[-4.5mm]
    {\large {{Testing OpenAI's models on non-public book content }}}\\[0.3cm]
}

\date{}

\author[1]{Sruly Rosenblat}
\author[1,2]{Tim O'Reilly}
\author[1,3,4]{Ilan Strauss}

\affil[1]{AI Disclosures Project, Code for Science \& Society (CS\&S), Portland, USA}
\affil[2]{O'Reilly Media, Sebastopol, USA}
\affil[3]{South African Research Chair in Industrial Development (SARChI-ID), University of Johannesburg, Johannesburg, South Africa}
\affil[4]{Institute for Innovation and Public Purpose, The Bartlett School of Architecture, University College London, London, UK}
\affil[ ]{Ilan Strauss: \href{mailto:ilan@aidisclosures.org}{ilan@aidisclosures.org}\\Sruly Rosenblat: \href{mailto:sruly@aidisclosures.org}{sruly@aidisclosures.org}}

\begin{document}

	\maketitle
\vspace*{-2.5em}

	\begin{abstract}
		\setstretch{1.15}

\thispagestyle{empty}

Using a legally obtained dataset of 34 copyrighted O'Reilly Media books (13,962 paragraphs), we apply the DE-COP membership inference attack to test whether OpenAI models show recognition of non-public (pay-walled) content they were not licensed to train on. Our main contribution is examining non-public books, which allows us to consider
potential access violations and not just training inclusion. Our results suggest that GPT-4o, OpenAI's more recent and capable model in our sample, exhibits patterns consistent with recognition of non-public books, with an AUROC score of 0.82 (95\% bootstrapped CI: 0.60--0.96) -- where above 50\% is better than random chance. GPT-3.5 Turbo (trained two years prior) and the smaller GPT-4o Mini both score near chance recognition on non-public data (0.54 and 0.56, respectively). Results are robust to independent paraphrasers (Gemma and Qwen) and leave-one-book-out sensitivity analyses. We also test both GPT-4o and GPT-4o mini, with the same October 2023 training cutoff date, to ensure the model is not simply identifying language shifts over time. These findings support disclosure requirements for pre-training data provenance and formal licensing markets for training content. Results remain preliminary given the small book sample, possible contamination of post-cutoff controls, and differences across models in size, architecture, and training composition.\\

\noindent \textit{Keywords}: Membership Inference Attacks, Large Language Models, Copyright Issues, Data Access Violations, Pre-Training Data, Architecture of Participation.

	\end{abstract}

	\clearpage
	\setcounter{page}{0}

	\newpage

\setstretch{1.5}

\section{Introduction: Investigating potential access violations}
\label{sec:intro}
\setcounter{page}{1}
Large Language Models (LLMs) require incredible amounts of public and non-public data to learn human language (called the `pre-training' stage). Yet the origins and legal status of this pre-training data remains largely undisclosed by the corporations that gather and use it \parencite{openai2023gpt4card, anthropic2023claude3}. Several high-profile legal proceedings indicate that major AI companies may train on non-public, often illegally obtained, content \parencite{nyt2023complaint, roth2024pile, arstechnica2025meta}.

\textit{This paper examines whether non-publicly accessible (non-public) copyrighted O'Reilly Media books were included in the training datasets of OpenAI's GPT series of models}. Each O'Reilly Media book contains both publicly accessible, free-to-use preview content, and non-public, pay-walled content. This allows us to investigate separately whether the models were primarily trained on publicly available data or if their datasets potentially circumvented paywall restrictions and used non-public data (Figure \ref{fig:model_training_split}).

\textit{We employ the DE-COP membership inference attack by \textcite{decop2024}} to test whether a model can reliably differentiate between human-authored (O'Reilly Media) texts and paraphrased LLM versions of the text that we generate. If it can, then the model might have prior knowledge of the text from its training \parencite{decop2024}. By systematically probing a model's knowledge of texts published before and after its training cutoff date, we can estimate the probability of particular book extracts having been included in a model's training data (DE-COP measures paragraph-level recognition; AUROC then aggregates these scores to assess overall separability between potentially-seen and unseen books).

We test OpenAI's GPT-3.5 Turbo, GPT-4o Mini, and GPT-4o models across 13,962 paragraphs from 34 O'Reilly books for potential access violations, distinguishing between public and non-public content extracted from the same books. Throughout the paper, we refer to higher DE-COP scores on pre-cutoff books as ``recognition''; it does not, by itself, establish training inclusion but is evidence for it (Section \ref{sec:method}). On the basis of AUROC scores calculated for each GPT Model (using 24 books for GPT-4o and GPT-4o Mini, and 26 for GPT-3.5 Turbo), where 50\% reflects no detectable recognition by the model, \textbf{we find that}:

\begin{enumerate}
\item \textit{Recognition of non-public data is markedly stronger in OpenAI's newer model, a pattern consistent with an increased role of non-public data over time}. GPT-4o achieves an AUROC score of 0.82 (95\% bootstrapped CI: 0.60--0.96), while GPT-3.5 Turbo, with a training cutoff two years prior, scores just above 0.50. This difference may reflect changes in training data composition, but could also be driven by differences in model size or architecture.

\item \textit{GPT-4o shows suggestively stronger recognition of non-public O'Reilly book content than of publicly accessible samples}, with AUROC scores of 0.82 (CI: 0.60--0.96) for non-public data vs 0.64 (CI: 0.36-0.93) for public data. This public/non-public comparison is statistically significant under a clustered paired book-level bootstrap (see \ref{sec:bootstrap-diff}), but matched local controls do not reproduce a statistically significant private-text advantage (see \ref{sec:public-private-controls}). We therefore treat the public/non-public gap as secondary and descriptive, not as causal evidence that paywall status itself drives the effect.

\item \textit{GPT-4o Mini shows little recognition of O'Reilly data}. We find that GPT-4o Mini, with the same training cutoff as GPT-4o, shows little to no recognition of O'Reilly data -- public or non-public. This may reflect reduced memorization capacity in a smaller model rather than differences in training data, or some combination of both \parencite{meeus2024copyright, morris2025much}.
\end{enumerate}

If access violations occurred, they might have occurred via the LibGen database, as all of the O'Reilly books tested were found in it. Alternatively data may have come from Books3 \parencite{jia2025cloze} or from snippets distributed throughout the internet. No licensing agreement existed between OpenAI and O'Reilly Media at the time of this study.

As a robustness check, we show that although newer LLMs have an improved ability to distinguish human-authored from machine-generated language \textit{regardless of whether a particular text was trained on}, this does not reduce the method's ability to classify data as being seen or not.

Our study design accounts for the potential of time-specific differences in language to bias our results \parencite{duan2024membership, das2024blind}, which can arise because we split our sample (of potentially trained on and so in-sample, vs. not trained on and so out-of-sample) by date. Such bias can occur if the DE-COP test mistakes language that the model is simply ``familiar'' with (due to temporal shifts) for content the model was trained on. To ensure that this bias does not drive our findings, we test two models (GPT-4o and GPT-4o Mini) that were both trained on data from the same period. Because these two models show notably different results, time-specific effects are unlikely to be the determining factor (although differences in model size and architecture may obscure differences). We also conduct multiple robustness tests in the appendix \ref{sec:robustness-appendix} to ensure our findings are not driven by the model size or architecture.

Our study contributes to research on detecting unauthorized data usage in AI training \parencite{neighbor2023, zhang2024membership} by applying membership inference methods to legally sourced non-public copyrighted material. Unlike earlier studies that primarily use publicly available datasets \parencite{shi2023detecting, decop2024, duan2024membership}, our public/non-public split within the same books enables the examination of how the paywall status of text affects model recognition.

Our findings highlight the need for stronger accountability in AI companies' model pre-training process. Liability provisions that incentivize improved corporate transparency in disclosing data provenance \parencite{oreilly2024aisoriginalsin} may be an important step to facilitating commercial markets for training data licensing and remuneration \parencite{ft_2025_copyright_wars}. Membership inference attacks can help pressure model developers to negotiate such agreements, but by themselves are insufficient, especially given their limited efficacy against smaller models, more advanced models, and models with certain post-training features \parencite{balaji2024fairuse, satvaty2024undesirable, ft_2025_copyright_wars}.

By way of robustness, our results are based on a small sample of books and so are potentially sensitive to individual book results. It is also difficult to isolate the role of model size on our results and is an important area for future research. We ran several robustness tests in Appendix \ref{sec:robustness-appendix}.

Section \ref{sec:dm} outlines our books dataset and our implementation of the DE-COP method. Section \ref{sec:findings} presents our findings. Section \ref{sec:discussion} discusses their policy implications for establishing formal commercial markets for content creator training data. Appendix \ref{sec:appendix} contains more details on our sample and analysis.

\section{Data and Methods}
\label{sec:dm}
This section first details our O'Reilly dataset of 34 books and explains how its division into publicly accessible vs non-public (effectively pay-walled) book samples enables us to look beyond public data to potentially identify access violations. Finally, we describe our research, which involves first testing the model's recognition of paragraphs from O'Reilly Media books, using the DE-COP membership inference attack method.

\subsection{Data: Public vs. non-public book data}
\label{sec:data}
Our dataset contains 34 copyrighted O'Reilly Media books lent to us, that we then split into a total of 13,962 paragraphs. Paragraphs are used to calculate the initial mean DE-COP score, one for each book, from which a single AUROC Score is then calculated across all books for each of OpenAI's models.

\begin{figure}[!ht]
  \centering
  \caption{\centering {\large{We split our sample of O'Reilly books by time period \& accessibility.}}}
  \vspace{2mm}
  \includegraphics[width=0.55\linewidth]{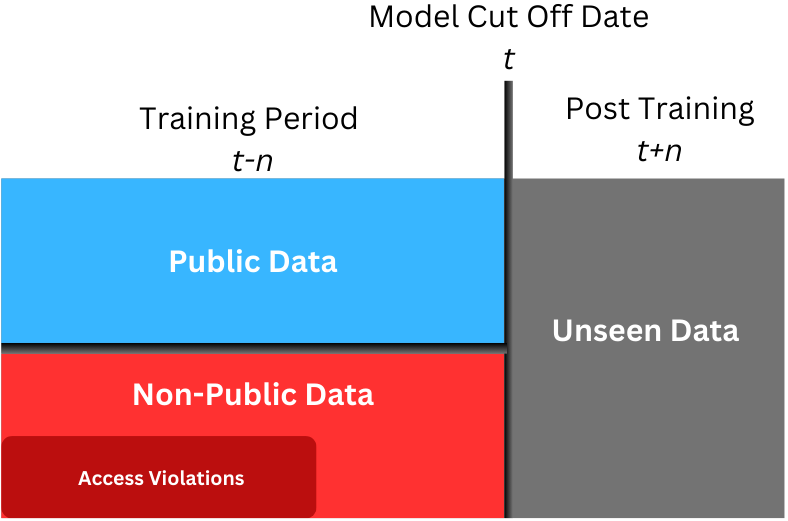}
  \vspace{3mm}
  \caption*{Note: Data published prior to a model's training completion ($t-n$) may have been trained on. Data published after a model's training cutoff ($t+n$) is known to not be in the model's training data (see \ref{sec:partially-seen} for where this may not be the case). Any portion of non-public data found to be included in a model's training would constitute an access violation (bottom left square).}
  \label{fig:model_training_split}
    \vspace{-4mm}
\end{figure}

The O'Reilly Media books dataset has the unique quality of containing both non-public (behind a paywall) and public (freely available) text within the same book. This allows us to differentiate between instances where a model was trained exclusively on public data and cases where potential access violations may have occurred. We define public text as any content made available by O'Reilly Media for content previews -- specifically the first 1,500 characters of each chapter as well as the entirety of chapters one and four. All other O'Reilly text we define as non-public.\footnote{This isn't an exact split, select paragraphs may have been copied to public articles, that would usually fall under fair use.}

To accurately measure the performance of the DE-COP membership inference attack method (discussed below), paragraph samples must be divided into two distinct categories, that in practice we can only approximate: data known to be included in the model's pre-training dataset and those known to be excluded. In our case we designate books published before the model's training cutoff ($t-n$) as \textit{possibly} in-dataset (previously seen and trained on) samples, and books published after a model's training cutoff ($t+n$) as \textit{known} out-of-dataset samples that the model could not have been trained on (see Figure \ref{fig:model_training_split}). ``Potential access violations'' are defined as the subset of non-public book paragraphs, published during the model's training period, for which the model shows recognition patterns consistent with training exposure.

We categorize books published before October 2023 (for GPT-4o and GPT-4o Mini), and before September 2021 (for GPT-3.5 Turbo) as \textit{potentially in-dataset} ($t-n$), and those books published after the model's training cutoff date as \textit{out-of-dataset} ($t+n$), where $t$ is the model training cutoff date (October 2023 and September 2021, respectively). This date is defined by the model developer as the last date that the model's pre-training dataset contains data for.

Our method of splitting our sample between potentially-in-dataset ($t-n$) and known out-of-dataset ($t+n$) \textit{by date} may introduce ``temporal bias'' into our findings \parencite{duan2024membership, das2024blind}, and in turn provide us with misleadingly high AUROC scores. This occurs when features in the data changes over time, creating distinguishable patterns between training and testing datasets split by time periods. Data then can be separately identified by an LLM based solely on the language varying with time -- in our case into potentially-in-dataset ($t-n$) and known out-of-dataset ($t+n$) data -- with no actual prior knowledge of the text itself.\footnote{Temporal bias is when the ability to infer membership through DE-COP, or any related method, is confounded by time-dependent changes in the data, rather than by genuine evidence that a particular example was (or was not) in the training set. Similarly, stylistic bias captures biases that arise from shifts in how data ``looks'' or is distributed (e.g., changes in vocabulary, writing style, or domain). Both these biases can appear if one naively splits data by time period, for instance, using older data for training and newer data for testing, without making any associated adjustments. In other words, during the DE-COP test the model might mistake familiar vs. unfamiliar language, for familiar vs. unfamiliar content they were trained on.}

To help account for this we test two different GPT models (GPT-4o and GPT-4o Mini) that were trained during the same period, and ideally on the same data, such that if our tests show very different AUROC results then temporal bias is unlikely to be the main driver. It is still possible that the models were trained on different datasets or have radically different architectures, the datasets and model architecture were never publicly disclosed to the best of our knowledge. We did additional controls for this (see \ref{sec:placebo}).

Our study design also helps isolate prior model knowledge of the data. Specifically, our results are unlikely due to GPT-4o simply being better at distinguishing human-authored from AI-generated text than GPT-4o Mini, as AUROC here measures the difference in ``recognition'' within the same model between books published prior to and after training completion. Even assuming a model has perfect identification capability, if it had not been trained on any of the test samples, we would expect an AUROC score of approximately 50\%.

It is also unclear how temporal bias would apply to DE-COP. As it is calculated based on how well the model identifies real text from paraphrases based in the same time period, with the same names, dates and concepts. For temporal bias to apply there would have to be some reason why paraphrases generated on text published prior to a models cutoff are more detectable than those generated on text published after besides for familiarity with the text. But with DE-COP we are not comparing paragraphs across years with different concepts or names. Instead, we are only directly comparing a paragraph with its paraphrased alternatives.

We attempted to carefully filter the dataset to avoid any ambiguous cases -- such as second edition books with potentially minor changes on the previous editions published during the training period -- that risk contaminating our ``unseen'' classification. However we discovered that two books with the unseen classification were tainted as discussed here \ref{sec:partially-seen}. To minimize edge cases where publication dates might overlap with training cutoffs, we excluded books published during a model's cutoff year from tests involving that specific model. Consequently, not every model was tested on precisely the same set of books: all AUROC computations use 24 books for GPT-4o and GPT-4o Mini (excluding 2023 publications and the two calibration books) and 26 books for GPT-3.5 Turbo (excluding 2021 publications and the two calibration books). See Appendix \ref{sec:book-details} for the full list.

\subsection{Methods: DE-COP and AUROC}
\label{sec:method}
We first use the \textbf{DE-COP} membership inference attack \parencite{decop2024} to determine whether a particular paragraph was part of a target model's training set. This works by quizzing an LLM with a multiple-choice test containing four options: one original human-authored O'Reilly book paragraph and three machine-generated paraphrases of that paragraph. To eliminate position bias (i.e., the tendency for models to favor certain answer positions such as ``A''), each paragraph is tested across all 24 possible permutations of the four options ($4! = 24$). A paragraph's ``guess rate'' is the proportion of these 24 permutations in which the model correctly identifies the original human-authored text. This process is repeated for nearly every paragraph in each of the 34 O'Reilly books tested, where a ``paragraph'' refers to adjacent sentences of approximately 120 words (not necessarily corresponding to paragraph breaks in the source material).\footnote{See Appendix \ref{sec:prompts_and_settings_openai} for an example of the prompt format used.}

If the model frequently identifies the original text for books published during its training period, this likely indicates prior recognition of that content. The key comparison is between this guess rate on books published before the model's training cutoff and the baseline guess rate on books published after the cutoff -- content the model could not have been trained on.

We test GPT-3.5 Turbo, GPT-4o Mini, and GPT-4o in this manner and use Claude 3.5 Sonnet\footnote{This differs from the paraphrase model used in \textcite{decop2024}. We chose Claude 3.5 Sonnet as it was the most powerful model from Anthropic at the time of testing.} to generate the paraphrased, machine versions, of the text. We decided to focus on OpenAI's models as at the time of testing they provided top 20 log probabilities making the method more interpretable. Table \ref{tab:sample_sizes} shows the paragraph sample sizes used to calculate the DE-COP scores for each paragraph.\\[-2mm]

\begin{table}[ht!]
\centering
  \caption{\centering {\large{DE-COP -- Paragraph-Level Sample Size and Average Word Count by Model (by type)}}}
  \vspace{-1mm}
\begin{tabular}{llrr}
\toprule
Model & Data-Split & Sample Size (n) & Average Word Count \\
\midrule
\multirow{4}{*}{GPT-4o}
  & Public                  & 1,965 & 112 \\
  & Non-Public              & 8,997 & 113 \\
  & Potentially In-Dataset  & 8,985 & 113 \\
  & Out-of-Dataset          & 1,977 & 110 \\
\midrule
\multirow{4}{*}{GPT-4o Mini}
  & Public                  & 1,968 & 112 \\
  & Non-Public              & 9,005 & 113 \\
  & Potentially In-Dataset  & 8,991 & 113 \\
  & Out-of-Dataset          & 1,982 & 110 \\
\midrule
\multirow{4}{*}{GPT-3.5 Turbo}
  & Public                  & 1,929 & 113 \\
  & Non-Public              & 6,171 & 113 \\
  & Potentially In-Dataset  & 2,084 & 114 \\
  & Out-of-Dataset          & 6,016 & 113 \\
\bottomrule
\end{tabular}

\vspace{2mm}
\caption*{Note: Sample sizes (in paragraphs) and average word counts across different data splits for each model. Potentially in-dataset represents data published prior to a model's cutoff date; out-of-dataset represents data published afterward.}
\label{tab:sample_sizes}
\end{table}

The second step in our study is to use the DE-COP quiz scores or guess rates' generated above to calculate \textbf{AUROC Scores} (Area Under the Receiver Operating Characteristic). Where DE-COP measures a model's ability to identify original human-authored text at the paragraph level, AUROC aggregates these paragraph-level scores to evaluate whether there is a meaningful difference (or separability') between how a model handles content that it was potentially trained on versus content published after its training was completed. AUROC measures a classifier's ability to distinguish between two classes, with scores ranging from 0 to 1, with 0.5 representing random chance and values closer to 1 indicating a strong ability to accurately `discriminate' (i.e., classify) between the two classes (or categories).

In our case, AUROC measures the ability to separate books that may have been trained on ($t-n$), from books the model could not have seen ($t+n$). A high AUROC score, therefore, indicates that the model's recognition scores systematically separate pre-cutoff from post-cutoff books -- a pattern consistent with training exposure to many of the pre-cutoff books. The threshold that optimally separates the two classes is determined by the AUROC calculation itself (i.e., the point on the ROC curve that maximizes the true positive rate while minimizing the false positive rate); we describe the specific thresholding variants used in Appendix \ref{sec:AUROC-scores}.

We calculate AUROC scores at both the paragraph and book levels, though our primary finding is at the book level (Table \ref{tab:gpt_paragraphs}). AUROC scores are calculated on the book-level sample sizes: being 24 for GPT-4o, 24 for GPT-4o Mini, and 26 for GPT-3.5 Turbo. In summary, DE-COP produces a paragraph-level guess rate, which is averaged to the book level, and AUROC then measures separability between potentially-seen and unseen books based on these scores.\\[-2mm]

\begin{table}[ht!]
\centering
\renewcommand{\arraystretch}{1.5}

  \caption{\centering {\large{AUROC Sample - Paragraph and Book Sample Sizes by Model}}}
  \vspace{-1mm}
\begin{tabular}{lrrrr}
\toprule
Model & Total Paragraphs & Non-Public & Public & Books \\
\midrule
GPT-4o & 10,962 & 8,997 & 1,965 & 24 \\
GPT-4o Mini & 10,973 & 9,005 & 1,968 & 24 \\
GPT-3.5 Turbo & 8,100 & 6,171 & 1,929 & 26 \\
\bottomrule
\end{tabular}

\vspace{2mm}
\caption*{Note: For GPT-4o, we use a sample of 10,962 paragraphs across 24 books, of which 8,997 are non-public and 1,965 are public. Similarly, for GPT-4o Mini we use 10,973 paragraphs (9,005 non-public and 1,968 public) across 24 books. Finally, GPT-3.5 Turbo used 8,100 paragraphs, with 6,171 non-public and 1,929 public paragraphs across 26 books. Counts in a previous version of this table mistakenly included the two calibration books (349 paragraphs for GPT-3.5 Turbo), which are excluded from all AUROC computations; counts now match the analysis sample exactly.}
\label{tab:gpt_paragraphs}
\end{table}

\section{Findings}
\label{sec:findings}
We present our core findings below, based on book level AUROC scores. We first calculate DE-COP guess rates for public and non-public book paragraphs within each book. Next, we calculate the mean DE-COP guess rate for each book based on these paragraphs, and use this to calculate an AUROC score for each large language model pooled across books. We run and test the various LLMs via Python (Google Colab) using OpenAI and Anthropic's batch API (Appendix \ref{sec:prompts_and_settings_openai} and \ref{sec:prompts_and_settings_claude}).

In what follows an AUROC score of 0.50 indicates no detectable recognition by the model; while test scores approaching 1.0 suggest near-perfect classification ability (between potentially in-dataset and out-of-dataset samples) -- based on the previously estimated DE-COP guess rate. Our confidence interval are calculated using the bootstrap method.\\[-2mm]

\begin{figure}[H]
  \centering
  \caption{\centering {\large{AUROC Scores: Recognition Patterns Consistent with Pre-Training Exposure}}}
    \vspace{-2mm}
  \includegraphics[width=0.82\linewidth]{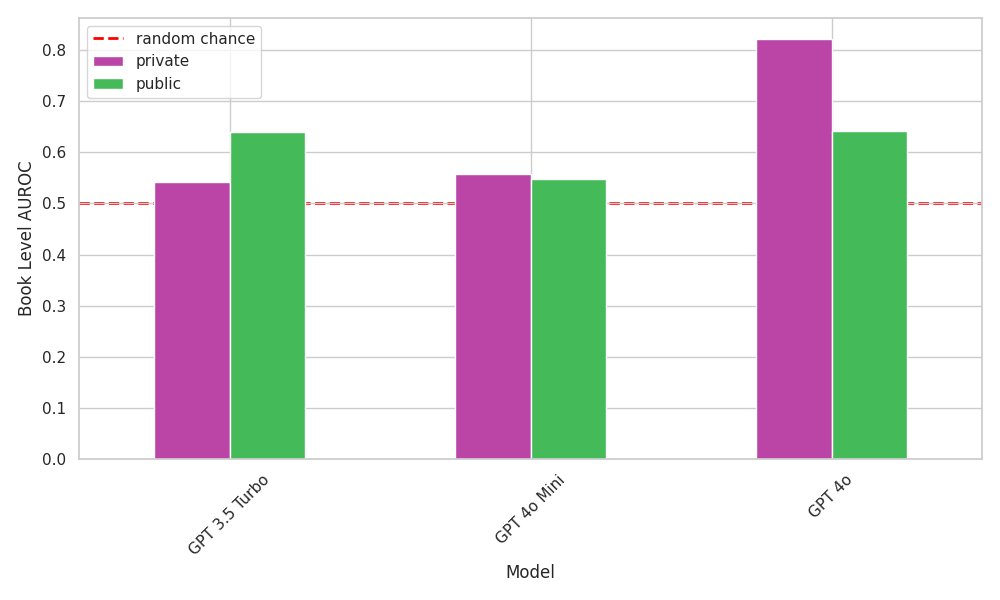}
  \caption*{Note: Showing book level AUROC scores (n = 24 for GPT-4o and GPT-4o Mini for both public and non-public; n = 26 for GPT-3.5 Turbo for both public and non-public) across models and data splits (see Table \ref{tab:sample_sizes} for sample sizes). Book level AUROC is calculated by averaging the identification rates of all paragraphs within each book and running AUROC on that. Bootstrapped 95\% confidence intervals are reported in Table \ref{tab:books_confidence_interval}.}
  \label{fig:auroc_book_lvl}
\end{figure}
\vspace{-4mm}

\textit{We find that OpenAI's more recent and capable model shows markedly stronger recognition of non-public O'Reilly book content than its older model}. Figure \ref{fig:auroc_book_lvl} shows that GPT-4o strongly recognizes pay-walled O'Reilly book content (book level AUROC 0.82), while OpenAI's GPT-3.5 Turbo, with a training cutoff two years prior in September 2021, shows no recognition (AUROC just above 0.50). This reflects a notably improved ability to distinguish between non-public books that were potentially included in the training dataset and those published after the model's pre-training cutoff. This score suggests recognition of many non-public O'Reilly books published prior to GPT-4o's training cutoff date (October 2023), raising the possibility that such books were included in its training data.

Secondly, Figure \ref{fig:auroc_book_lvl} also shows that \textit{GPT-4o exhibits somewhat stronger recognition of non-public O'Reilly book content compared to publicly accessible samples}, with AUROC scores of 0.82 (non-public) vs 0.64 (public). We treat this public/non-public contrast as suggestive and secondary, since matched local controls leave stylistic or structural differences between preview and non-preview text as a plausible explanation (Appendix \ref{sec:public-private-controls}). The core result is therefore not that paywall status alone explains recognition, but that GPT-4o's recognition is elevated within non-public text when comparing pre-cutoff books to post-cutoff books.

 Our book level AUROC scores calculated on the mean DE-COP scores for each book were often significantly higher than AUROC done on the paragraph level (Appendix Figure \ref{fig:auroc_variations}). Our AUROC results, being much starker at the book level compared to the paragraph level. This difference is similar to \textcite{puerto2024scaling}, who finds that aggregating results over larger data units significantly enhances the performance of membership inference attacks. However, it is important to note our confidence interval was also much larger at the book level because of there being a small sample size -- under 30 books and only 5 post-cutoff books for GPT-4o and GPT-4o Mini.

GPT-4o's seemingly high recognition of O'Reilly Media books could reflect the deliberate presence of these books in its training data. However, some of this familiarity may have been acquired through more benign means -- for example, excerpts from these books may have entered the dataset via user queries or appear in fair use quotations throughout the internet.\\[-2mm]

\subsection{Robustness and limitations}
\label{sec:robustness}

One limitation of our study is that smaller models may be harder to test accurately in membership inference attacks and as such the gap between models may come down to memory and not training. However, prior research has established no clear relationship between model size and model memorization, noting that larger models can memorize more samples (approximately 3.6 bits per parameter), but also that membership inference becomes harder on larger models as the datasets they are trained on also grows considerably \parencite{morris2025much}: ``bigger models can memorize more samples, and making [sic] datasets bigger makes membership inference harder.'' Whatever the case, GPT-4o Mini\footnote{OpenAI does not disclose model sizes but GPT-4o Mini is smaller than GPT-4o and presumably smaller than GPT-3.5 Turbo.} recorded AUROC scores of 0.55 on public data and 0.56 on non-public data, both near random chance.

\begin{figure}[H]
  \centering
  \caption{\centering {\large{DE-COP Guess Rate Improves:} \textnormal{More capable models identify human text even when not trained on it.}}}
   \vspace{-2mm}
\includegraphics[width=0.82\linewidth, trim={0 1cm 0 0}, clip]{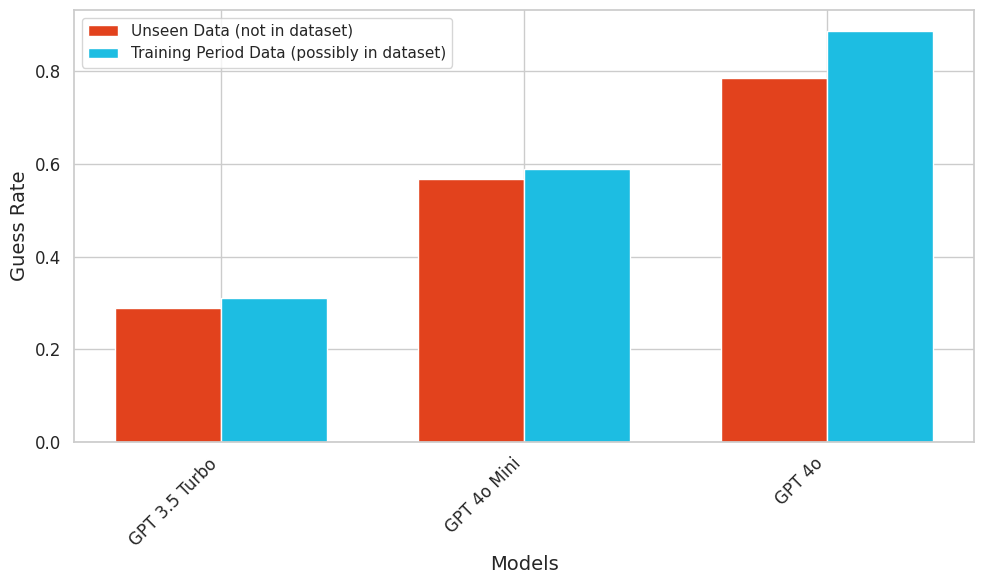}
  \vspace{2mm}
  \caption*{Note: DE-COP guess rates (i.e., identification rates) pooled across all books for OpenAI models. Red bars represent unseen data published after a model finished training ($t+n$), and blue bars represent data published before the cutoff date ($t-n$) that is suspected to be in the training set. See Figure \ref{fig:model_training_split} for more. This should not be confused for AUROC scores, AUROC is calculated on these raw DE-COP scores by measuring the separability of DE-COP scores between data splits.}
  \label{fig:guess_rates}
\end{figure}

Second, we note that improving LLM capabilities can make the identification of pre-training data through membership inference attacks more difficult. As per Figure \ref{fig:guess_rates}, we find that OpenAI's models' ability to correctly identify human-authored text, among paraphrased LLM alternatives, improves with model capabilities, \textit{even for texts the model could not have been trained on} -- meaning those texts published after the model's training cutoff. Figure \ref{fig:guess_rates} shows the baseline DE-COP identification rate on books published after the model's training cutoff (unseen books). This increased from 0.31 for GPT-3.5 Turbo (training finished September 2021), to 0.57 for GPT-4o Mini (training finished October 2023), and to 0.78 for GPT-4o (training finished October 2023).

Once the baseline guess rate (`identification rate') exceeds 96\%, the difference between potentially in-dataset and out-of-dataset paragraphs could become undetectable at the paragraph level. For now, however, the gap remains sufficiently large to reliably separate the two categories when calculating AUROC score, particularly when aggregating results at the book level.

\vspace{3mm}
\vspace{-4mm}

We also performed several additional robustness checks, which are reported in full in Appendix \ref{sec:robustness-appendix}. These include a clustered paired bootstrap of the public/non-public difference (Appendix \ref{sec:bootstrap-diff}), tests with alternative paraphrasing models (Appendix \ref{sec:paraphraser-robustness}), placebo and negative control tests (Appendix \ref{sec:placebo}), a leave-one-book-out sensitivity analysis (Appendix \ref{sec:lobo}), and matched public/private style and position controls (Appendix \ref{sec:public-private-controls}). These checks support the main GPT-4o non-public recognition result while limiting how strongly the public/non-public contrast should be interpreted.

\section{Discussion: Towards functional content AI marketplaces?}
\label{sec:discussion}

Although the evidence presented here on potential access violations is specific to OpenAI and O'Reilly Media books, and based on a small sample with wide confidence intervals, similar dynamics may exist across AI model developers. Our findings aim to motivate greater transparency in data collection and usage practices.

Our findings, alongside similar studies \parencite{ahmed2026extracting, arstechnica2025meta} suggest that current AI model development practices may be creating what \textcite{oreilly2024aisoriginalsin} describes as an ``extractive dead end'', creating not just a legal challenge but an existential one for the Internet's content ecosystem. The economic implications of uncompensated training data usage extend beyond individual copyright holders to the broader sustainability of professional content creation. If AI companies extract value from a content creator's produced materials without fairly compensating the creator, they risk depleting the very resources upon which their AI systems depend \parencite{oreilly2024aisoriginalsin}.

This dynamic creates a tragedy of the commons.\footnote{As \textcite{longpre2024consent} notes: ``in less than a year, $\sim5\%$ of the tokens in C4 and other major corpora have recently become restricted by robots.txt. And nearly 45\% of these tokens now carry some form of restrictions from the domain's Terms of Service.''} If left unaddressed, uncompensated training data could lead to a downward spiral in the Internet's content quality and diversity. As revenue streams for professional content creation diminish, fewer resources will be dedicated to producing the high-quality, accurate, and diverse human content that AI systems rely on for training -- and inference.

Our key finding, that OpenAI may have  trained their GPT-4o model on non-public data, is only preliminary and is based on a small sample of books and subject to the above methodological caveats. Membership inference attacks of a model's outputs are not a substitute for detailed -- ideally programmatic -- model cards that disclose and disaggregate the sources of model training data \parencite{mitchell2019model, gebru2021datasheets}. However, requiring smaller companies to sift through their pre-training dataset and individually identify the sources for each of their training inputs is unrealistic without tools and standards designed for this purpose.

Common Corpus \parencite{langlais2024releasing}, a large pre-vetted training dataset, is one way around this issue. By centralizing the data cleaning process and providing verifiable pre-training data as a common public good, datasets like Common Corpus could enable smaller firms to train models on non-proprietary data, and easily facilitate disclosure \parencite{langlais2024releasing}. Specialized data auditing companies are already arising but limited in what they can achieve without specific standards.

Ensuring that IP holders know when their work has been used in model training represents a crucial first step toward establishing AI markets for content creator data. Technical methods for this are still in their infancy \parencite{grosse2023studying, minkplus2024}. But when applied to specific types of content, such as music, these methods seem to achieve better results, with at least one new music platform already apparently being able to attribute AI generated music outputs to specific music training inputs \parencite{musicalAI2025}.

More broadly, given the apparent importance of high-quality paywalled content for model training, structured markets for licensing such data remain both feasible and necessary. If the current lack of transparency around training data provenance persists, it could harm both content creators and AI developers. Content creators lose revenue and incentive to produce the high-quality material that models depend on, while developers face reputational and legal risk alongside a potential degradation in the quality of available training content -- particularly as a growing share of online content becomes AI-generated \parencite{oreilly2024aisoriginalsin}. Liability regimes and disclosure requirements may be necessary to catalyze viable marketplaces for various types of model training and inference content \parencite{ft_2025_copyright_wars}.

\section{Conclusion}
\label{sec:conclusion}

This study applies the DE-COP membership inference attack to 34 legally obtained copyrighted O'Reilly Media books to examine whether OpenAI's models recognize non-public, paywalled content. GPT-4o shows strong recognition of non-public book content (book level AUROC 0.82), while GPT-4o Mini and GPT-3.5 Turbo do not, though the wide confidence intervals and small sample size warrant caution in interpretation. Our principal contribution is the application of membership inference methods to legally obtained non-public material, enabling the detection of potential access violations that studies using only public data cannot identify. Although our evidence is specific to OpenAI and O'Reilly Media, the underlying dynamics likely extend to other model developers and content publishers. Future work should expand the sample of books and publishers tested, explore complementary detection methods, and investigate how model size and architecture interact with membership inference performance.

\section*{Declarations}
\begingroup\setstretch{1.1}

\subsection*{Author Contributions}
S.R. did compute, statistical analysis and AUROC method, appendix, graphs, and tables. I.S. did paper write-up, structure, core findings, policy discussion. T.O. did topic conceptualization and research design (public vs. non-public data). All authors reviewed the manuscript.

\subsection*{Data Availability}
An anonymized version of our dataset with most text removed can be found at \url{https://drive.google.com/drive/folders/10A3FIe6hjb_pURaDULWSSJwUXS-gUSgG}. The original data has not been made public due to copyright concerns.

\subsection*{Ethics Declaration}
The paragraphs analyzed were taken from lawfully obtained copies of copyrighted books. We did not circumvent access controls. We released data on our results with most text content removed to allow others to review our findings without exposing copyrighted material. All code and numerical outputs needed to reproduce our results have been made available.

\subsection*{Funding Declaration}
We are grateful for funding received from the Alfred P. Sloan Foundation, Omidyar Network, Patrick J. McGovern Foundation, and the O'Reilly Foundation.

\subsection*{Conflicts of Interest}
One of the coauthors, Tim O'Reilly, is the CEO of O'Reilly Media. We additionally received part of our funding from the O'Reilly Foundation.
\par\endgroup

\newpage

\begin{singlespace}
\nocite{zhao2024explainability}
\printbibliography
\end{singlespace}
\cleardoublepage
\normalsize

\appendix
\section{Appendix}
\label{sec:appendix}

\subsection{Additional Details About Our Dataset}
\label{sec:book-details}

We tested OpenAI's models on a total of 34 books, but not all books were used for every model. The table below lists the books used and their publication dates. For each model, we excluded any data published in the year the model completed its training from our testing.\\[-2mm]

\captionsetup[longtable]{singlelinecheck=false}
\setlength\LTleft{0pt}
\setlength\LTright{0pt}
\setlength\LTcapwidth{14cm}

\footnotesize

\begin{longtable}{|C{4.1cm}|C{1.7cm}|C{2.0cm}|C{2.0cm}|C{2.0cm}|}
\caption{Detailed information about the books included in our dataset.}
\label{tab:book_details}\\
\hline
\textbf{Title} & \textbf{Date} & \textbf{GPT-3.5 Turbo Paragraph Count} & \textbf{GPT-4o Mini Paragraph Count} & \textbf{GPT-4o Paragraph Count} \\
\hline
\endfirsthead

\hline
\textbf{Title} & \textbf{Date} & \textbf{GPT-3.5 Turbo Paragraph Count} & \textbf{GPT-4o Mini Paragraph Count} & \textbf{GPT-4o Paragraph Count} \\
\hline
\endhead

\endlastfoot
97 Things Every Information Security Professional Should Know & 2021-09-14 & --- & 315 & 314 \\ \hline
AI-Powered Business Intelligence                   & 2022-06-10 & 239 & 397 & 396 \\ \hline
Advancing into Analytics                           & 2021-04-18 & --- & 157 & 157 \\ \hline
Applied Machine Learning and AI for Engineers      & 2022-11-10 & 329 & 353 & 353 \\ \hline
Azure Cookbook                                     & 2023-06-29 & 42 & --- & --- \\ \hline
Building Green Software                            & 2024-03-11 & 226 & 416 & 414 \\ \hline
Building Knowledge Graphs                          & 2023-06-26 & 160 & --- & --- \\ \hline
Building Recommendation Systems in Python and JAX  & 2023-12-11 & 311 & --- & --- \\ \hline
Building Solutions with the Microsoft Power Platform  & 2023-01-06 & 283 & --- & --- \\ \hline
C\# 8.0 in a Nutshell                              & 2020-05-12 & 335 & 335 & 334 \\ \hline
Cloud Native Go                                    & 2021-04-20 & --- & 358 & 358 \\ \hline
Communicating with Data                            & 2021-10-03 & --- & 446 & 446 \\ \hline
Continuous Deployment                              & 2024-07-25 & 584 & 584 & 582 \\ \hline
Data Quality Fundamentals                          & 2022-09-02 & 447 & 447 & 447 \\ \hline
Deciphering Data Architectures                     & 2024-02-07 & 363 & 477 & 477 \\ \hline
Delta Lake: Up and Running                         & 2023-10-17 & 187 & --- & --- \\ \hline
DevOps Tools for Java Developers                   & 2022-04-15 & 304 & 467 & 464 \\ \hline
Distributed Tracing in Practice                    & 2020-04-14 & 323 & 578 & 578 \\ \hline
FastAPI                                            & 2023-11-13 & 79 & --- & --- \\ \hline
Genomics in the Cloud                              & 2020-04-08 & 479 & 767 & 767 \\ \hline
Leading Lean                                       & 2020-01-23 & 301 & 486 & 486 \\ \hline
Learning Digital Identity                          & 2023-01-10 & 478 & --- & --- \\ \hline
Natural Language Processing with Spark NLP       & 2020-06-25 & 135 & 271 & 271 \\ \hline
Policy as Code                                     & 2024-07-09 & 235 & 335 & 334 \\ \hline
Practical Natural Language Processing            & 2020-06-17 & 292 & 410 & 410 \\ \hline
Programming C\# 10                                 & 2022-08-05 & 1059 & 1538 & 1538 \\ \hline
Prompt Engineering for Generative AI \textit{(excluded)}             & 2024-05-16 & 262 & 304 & 304 \\ \hline
RESTful Web API Patterns and Practices Cookbook    & 2022-10-17 & 276 & 444 & 444 \\ \hline
Scaling Machine Learning with Spark              & 2023-03-09 & 291 & --- & --- \\ \hline
Security and Microservice Architecture on AWS      & 2021-09-08 & --- & 452 & 452 \\ \hline
Software Architecture: The Hard Parts              & 2021-10-25 & --- & 404 & 404 \\ \hline
The Customer-Driven Culture: A Microsoft Story     & 2020-03-10 & 219 & 366 & 366 \\ \hline
Web API Cookbook \textit{(excluded)}                        & 2024-03-28 & 87 & 109 & 109 \\ \hline
Web Accessibility Cookbook                         & 2024-06-17 & 123 & 170 & 170 \\ \hline
\end{longtable}

\noindent\begin{minipage}{\linewidth}
\footnotesize
\noindent Note: Paragraph counts per model reflect the full quiz runs; the two calibration books and books published in a model's cutoff year are excluded from all AUROC computations, leaving 10,962 analysis paragraphs across 24 books for GPT-4o, 10,973 across 24 books for GPT-4o Mini, and 8,100 across 26 books for GPT-3.5 Turbo (see Table \ref{tab:gpt_paragraphs}).
\end{minipage}

\normalsize

In our study any books published in 2023 were excluded from tests involving GPT-4o and GPT-4o Mini, while books published in 2021 were omitted from any tests involving GPT-3.5 Turbo.

\begin{table}[H]
    \centering
    \caption{\centering {\large{There is a noticeable difference in phrasing between public and non-public text.}}}
    \begin{tabular}{|l|c|l|c|}
        \hline
        \multicolumn{2}{|c|}{\textbf{Public Split}} & \multicolumn{2}{c|}{\textbf{Non-Public Split}} \\
        \hline
        \multicolumn{1}{|c|}{Phrase} & \multicolumn{1}{c|}{Occurrences} & \multicolumn{1}{c|}{Phrase} & \multicolumn{1}{c|}{Occurrences} \\
        \hline
        in this chapter & 138 & as well as & 455 \\
        as well as      & 115 & one of the & 449 \\
        one of the      & 99 & be able to & 444 \\
        be able to      & 89 & the number of & 436 \\
        a lot of        & 85 & you want to & 399 \\
        \hline
    \end{tabular}
    \label{tab:phrase_frequency}
    \vspace{2mm}
    \caption*{Note: Shows most frequent phrases of each data split. The most common phrase in the public dataset introduces a chapter, likely because the public split primarily consists of the first 1,500 characters of each chapter.}
\end{table}

Table \ref{tab:phrase_frequency} displays the most common three-word phrases in the public and non-public datasets. These phrasing differences reflect that the public text is typically extracted from the first 1500 characters of each chapter (with the exception, being chapters one and four of each book where the entire chapter is public). The public portion of the dataset contains more language introducing a chapter as it mostly consists of the first 1,500 words of each chapter.

All paragraphs in \textit{Prompt Engineering for Generative AI} and  \textit{Web API Cookbook} were excluded from testing, originally to calibrate results. However, because we tested each paragraph across all 24 permutations of the four answer options (4!=24), position bias is already fully accounted for, making such calibration unnecessary. The two books remain excluded from all reported AUROC scores and statistical tests.

\subsection{AUROC Results}
\label{sec:AUROC-scores}

We found that there are various ways to calculate AUROC scores, some of which can lead to significantly different results (see Figure \ref{fig:auroc_variations} and Table \ref{tab:extra_performance_metrics}). For example, \textcite{decop2024} calculated their scores at the book level by first computing the mean guessing score across all paragraphs within each book. They then used these book level guessing rates to determine an optimal threshold, ultimately converting each book into a binary prediction based on whether it exceeded this threshold. This approach appeared to give a boost over doing the AUROC calculation directly without thresholding first (shown in Figure \ref{fig:auroc_variations}). We calculated AUROC scores using the following methods:

\begin{figure}[H]
  \centering
    \caption{\centering {\large{AUROC score is highly dependent on the data scale and method it is measured with.}}}
  \includegraphics[width=0.9\linewidth]{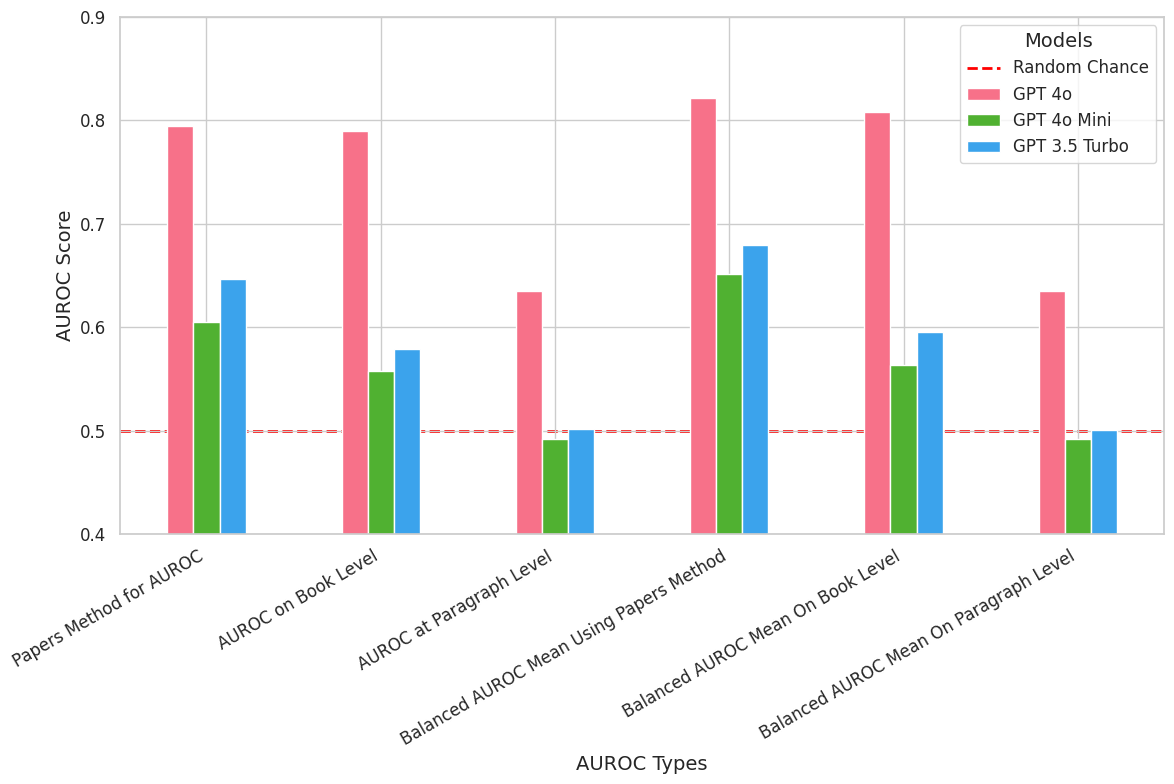}

  \label{fig:auroc_variations}
  \vspace{4mm}
  \caption*{Note: Despite measuring many different AUROC variations we always had a similar pattern of GPT-4o demonstrating the most knowledge, followed by GPT-3.5 Turbo and finally GPT-4o Mini showing the least recognition. See Tables:
  \ref{tab:sample_sizes}, \ref{tab:extra_performance_metrics}, \ref{tab:books_confidence_interval} and \ref{tab:chunk_confidence_interval}.
}
\end{figure}

\begin{itemize}
\item \textbf{Papers Method for AUROC} refers to the AUROC score approach from the DE-COP study \parencite{decop2024}. First, the optimal threshold is determined based on Youden's Index (TPR - FPR), \parencite{ibm_auc_spss}. Next, each book is given a binary value based on the optimal threshold. Finally, the AUROC score is computed on these binary predictions.
\item \textbf{Book Level AUROC} is calculated by averaging the identification rates across all paragraphs in a book and then computing the AUROC score using these averages.
\item \textbf{Paragraph Level AUROC} uses the identification rate for individual paragraphs to compute the AUROC score.
\item  \textbf{Balanced AUROC scores} are derived similarly to the other AUROC methods but are calculated from 100 subsets, each containing equal proportions of data from before and after the cutoff date. The mean scores from these subsets are then reported.
\end{itemize}

\begin{table}[H]
\centering
\caption{\large{All AUROC Metrics by Data Split and Model.}}
\begin{tabular}{lccc}
\toprule
 & \textbf{GPT-4o} & \textbf{GPT-4o Mini} & \textbf{GPT-3.5 Turbo} \\
\midrule
\multicolumn{4}{l}{\textbf{All Paragraphs}} \\
Papers Method for AUROC (Binary) & 0.79 & 0.61 & 0.65 \\
AUROC on Book Level & 0.79 & 0.56 & 0.58 \\
AUROC on Paragraph Level & 0.63 & 0.49 & 0.50 \\
Balanced AUROC Mean Using Papers Method & 0.82 & 0.65 & 0.68 \\
Balanced AUROC Mean on Book Level & 0.81 & 0.56 & 0.60 \\
Balanced AUROC Mean on Paragraph Level & 0.64 & 0.49 & 0.50 \\
\midrule
\multicolumn{4}{l}{\textbf{Public Paragraphs}} \\
Papers Method for AUROC (Binary) & 0.69 & 0.64 & 0.67 \\
AUROC on Book Level & 0.64 & 0.55 & 0.64 \\
AUROC on Paragraph Level & 0.60 & 0.48 & 0.51 \\
Balanced AUROC Mean Using Papers Method & 0.71 & 0.65 & 0.69 \\
Balanced AUROC Mean on Book Level & 0.64 & 0.53 & 0.63 \\
Balanced AUROC Mean on Paragraph Level & 0.60 & 0.48 & 0.51 \\
\midrule
\multicolumn{4}{l}{\textbf{Non-Public Paragraphs}} \\
Papers Method for AUROC (Binary) & 0.84 & 0.66 & 0.62 \\
AUROC on Book Level & 0.82 & 0.56 & 0.54 \\
AUROC on Paragraph Level & 0.64 & 0.50 & 0.50 \\
Balanced AUROC Mean Using Papers Method & 0.84 & 0.67 & 0.63 \\
Balanced AUROC Mean on Book Level & 0.82 & 0.56 & 0.54 \\
Balanced AUROC Mean on Paragraph Level & 0.64 & 0.50 & 0.50 \\
\bottomrule
\end{tabular}
\label{tab:extra_performance_metrics}
\vspace{2mm}
\caption*{Note: Shows all the AUROC scores that we calculated (see Table \ref{tab:sample_sizes} for sample sizes). Figure \ref{fig:auroc_variations} visualizes this table.}
\end{table}

Unless otherwise specified, the AUROC scores reported in this paper are book level AUROC. Throughout the paper we refer repeatedly to book level AUROC. However, when testing for robustness we found that the 95\% bootstrapped confidence intervals were very large at the book level (see Table \ref{tab:books_confidence_interval}).

\begin{table}[H]
\centering
  \caption{\centering {\large{Book Level AUROC Scores with Bootstrapped Confidence Intervals by Data Split.}}}
\begin{tabular}{lll}
\toprule
Model & Data-Split & Book Level AUROC \\
\midrule
\multirow{3}{*}{GPT-4o}
  & All     & 0.79 (0.53, 0.96) \\
  & Public  & 0.64 (0.36, 0.93) \\
  & Non-Public & 0.82 (0.60, 0.96) \\
\midrule
\multirow{3}{*}{GPT-4o Mini}
  & All     & 0.56 (0.25, 0.84) \\
  & Public  & 0.55 (0.20, 0.84) \\
  & Non-Public & 0.56 (0.28, 0.83) \\
\midrule
\multirow{3}{*}{GPT-3.5 Turbo}
  & All     & 0.58 (0.33, 0.83) \\
  & Public  & 0.64 (0.39, 0.86) \\
  & Non-Public & 0.54 (0.28, 0.77) \\
\bottomrule
\end{tabular}

\label{tab:books_confidence_interval}
\vspace{2mm}
\caption*{Note: We performed hierarchical bootstrapping using 1,000 bootstraps over all books not published in the year of a model's cutoff date (see Table \ref{tab:book_details}). To perform hierarchical bootstrapping, we repeatedly sampled random books and then random paragraphs within each book. See Table \ref{tab:sample_sizes} for sample sizes.}
\end{table}

This is likely attributable to our limited book count. We analyzed a sample of 34 books, each containing thousands of paragraphs. This small number of titles leads to a very wide bootstrapped confidence interval for the book level AUROC scores (see Table \ref{tab:books_confidence_interval}). Although not ideal, this outcome is expected if some books in our sample were part of the training data while others were not. Since we approximate the in-dataset and out-of-dataset groups with a small sample, any `mislabeled' data -- data that was assumed to be in-dataset but was not actually included or visa versa -- can disproportionately affect and skew the results.\\[-2mm]

\begin{table}[H]
\centering
  \caption{\centering {\large{Paragraph Level AUROC Scores with Bootstrapped Confidence Intervals by Data Split}}}
\begin{tabular}{lll}
\toprule
Model & Data-Split & Paragraph Level AUROC \\
\midrule
\multirow{3}{*}{GPT-4o}
  & All     & 0.63 (0.62, 0.65) \\
  & Public  & 0.60 (0.57, 0.63) \\
  & Non-Public & 0.64 (0.63, 0.66) \\
\midrule
\multirow{3}{*}{GPT-4o Mini}
  & All     & 0.49 (0.48, 0.51) \\
  & Public  & 0.48 (0.45, 0.51) \\
  & Non-Public & 0.50 (0.48, 0.51) \\
\midrule
\multirow{3}{*}{GPT-3.5 Turbo}
  & All     & 0.50 (0.49, 0.52) \\
  & Public  & 0.51 (0.48, 0.54) \\
  & Non-Public & 0.50 (0.48, 0.51) \\
\bottomrule
\end{tabular}
\vspace{2mm}
\label{tab:chunk_confidence_interval}
\caption*{Note: We performed bootstrapping using 1000 bootstraps over all paragraphs not published in the year of a model's cutoff date. See Table \ref{tab:sample_sizes} for sample sizes.}
\end{table}

In contrast, performing bootstrap at the paragraph level -- where data is pooled across books -- yields significantly smaller bootstrapped confidence intervals across all models (see Table \ref{tab:chunk_confidence_interval}), indicating that increasing the number of books would likely help tighten the book level bootstrapped confidence intervals.

\subsection{Robustness Analyses}
\label{sec:robustness-appendix}

This appendix reports the additional robustness checks referenced in Section \ref{sec:robustness}, together with their tables and figures.

\subsubsection{Clustered bootstrap of the public/non-public difference}
\label{sec:bootstrap-diff}

Our book level AUROC estimates are highly uncertain, with large bootstrapped confidence intervals. We tested the public/non-public difference with clustered paired bootstrap: books are resampled with replacement, both AUROC scores are computed on the same draw, and the difference is recorded (2,000 draws; 1,992 usable for GPT-4o and GPT-4o Mini after excluding draws that resampled only one side of the cutoff). For GPT-4o the non-public minus public difference was 0.18 (95\% CI: 0.02--0.38; two-sided p = 0.043), while GPT-4o Mini (0.02, CI $-0.19$--0.22, p = 0.87) and GPT-3.5 Turbo ($-0.10$, CI $-0.39$--0.14, p = 0.46) showed no difference (see Table \ref{tab:paired_bootstrap}).

\begin{table}[H]
\centering
\renewcommand{\arraystretch}{1.5}

\caption{\centering {\large{Clustered Paired Bootstrap of the Non-Public Minus Public Book Level AUROC Difference}}}
\begin{tabular}{lccccc}
\toprule
Model & Non-Public AUROC & Public AUROC & Difference & 95\% CI & P-value \\
\midrule
GPT-4o & 0.82 & 0.65 & 0.18 & (0.02, 0.38) & 0.04 \\
GPT-4o Mini & 0.56 & 0.55 & 0.02 & ($-0.19$, 0.22) & 0.87 \\
GPT-3.5 Turbo & 0.54 & 0.64 & $-0.10$ & ($-0.39$, 0.14) & 0.46 \\
\bottomrule
\end{tabular}
\caption*{Note: Books are resampled with replacement 2,000 times; the public and non-public book level AUROC scores are computed on each draw and their difference recorded (AUROC columns show means across draws). To obtain the two-sided p-value, we take the share of bootstrap draws in which the difference falls below zero and the share in which it falls above zero, and double the smaller of the two. }
\label{tab:paired_bootstrap}
\end{table}

\subsubsection{Paraphraser robustness}
\label{sec:paraphraser-robustness}

Because DE-COP scores depend on distinguishing original text from LLM paraphrases, our results could in principle reflect idiosyncrasies of the paraphrasing model (Claude 3.5 Sonnet) rather than the quizzed model's familiarity with the original text (even though we are comparing pre and post-cutoff De-Cop scores from the same paraphraser). To rule this out, we regenerated paraphrases with two additional models, Gemma 4 26B and Qwen 3.5 9B, whose model families are independent of both the original paraphraser and the quizzed models, using the identical prompt and temperature (Appendix \ref{sec:prompts_and_settings_claude}).

Owing to API cost, this check used a paired design on a stratified subsample of 1,280 paragraphs (40 per book; 302 public, 978 non-public) drawn from the paragraphs already tested in the main experiment, so AUROC scores are directly comparable on the same items. After the cutoff-year exclusion, 960 paragraphs enter each AUROC for GPT-4o and GPT-4o Mini, and 759 for GPT-3.5 Turbo, whose original run covered fewer snippets. On this subsample, GPT-4o's book level AUROC was 0.74 with the original Claude paraphrases, 0.85 with Gemma, and 0.81 with Qwen (see Table \ref{tab:paraphraser_robustness}). The ordering of models and the elevated recognition for GPT-4o were unchanged across paraphrasers, and AUROC scores with the independent paraphrasers were equal or higher for every target model. This indicates that the findings are not an artifact of a single paraphraser's style; in this paired subsample, the original Claude paraphrases yielded the lowest estimates.

\begin{table}[H]
\centering
\renewcommand{\arraystretch}{1.5}
\caption{\centering {\large{Paraphraser Robustness: Book Level AUROC by Paraphrasing Model (Paired Subsample)}}}
\begin{tabular}{lcccc}
\toprule
Target Model & Claude 3.5 Sonnet (original) & Gemma 4 26B & Qwen 3.5 9B & Paragraphs \\
\midrule
GPT-4o & 0.74 & 0.85 & 0.81 & 960 \\
GPT-4o Mini & 0.51 & 0.61 & 0.68 & 960 \\
GPT-3.5 Turbo & 0.43 & 0.55 & 0.55 & 759 \\
\bottomrule
\end{tabular}
\caption*{Note: Book level AUROC scores (all-paragraph split) on the stratified subsample of 1,280 paragraphs (40 per book), computed on the same paragraphs for every paraphraser so scores are directly comparable. The Claude column restricts the original run to the subsample.}
\label{tab:paraphraser_robustness}
\end{table}

\subsubsection{Temporal-boundary and negative-control tests}
\label{sec:placebo}

If our temporal split manufactured signal, for example through publication-year style drift, DE-COP would produce comparable separation at alternative date partitions.

\begin{figure}[H]
  \centering
  \caption{\centering {\large{For GPT-4o, the 95\% Bootstrapped Confidence Interval First Excludes Chance at the True Training Boundary}}}
  \vspace{-2mm}
  \includegraphics[width=0.95\linewidth]{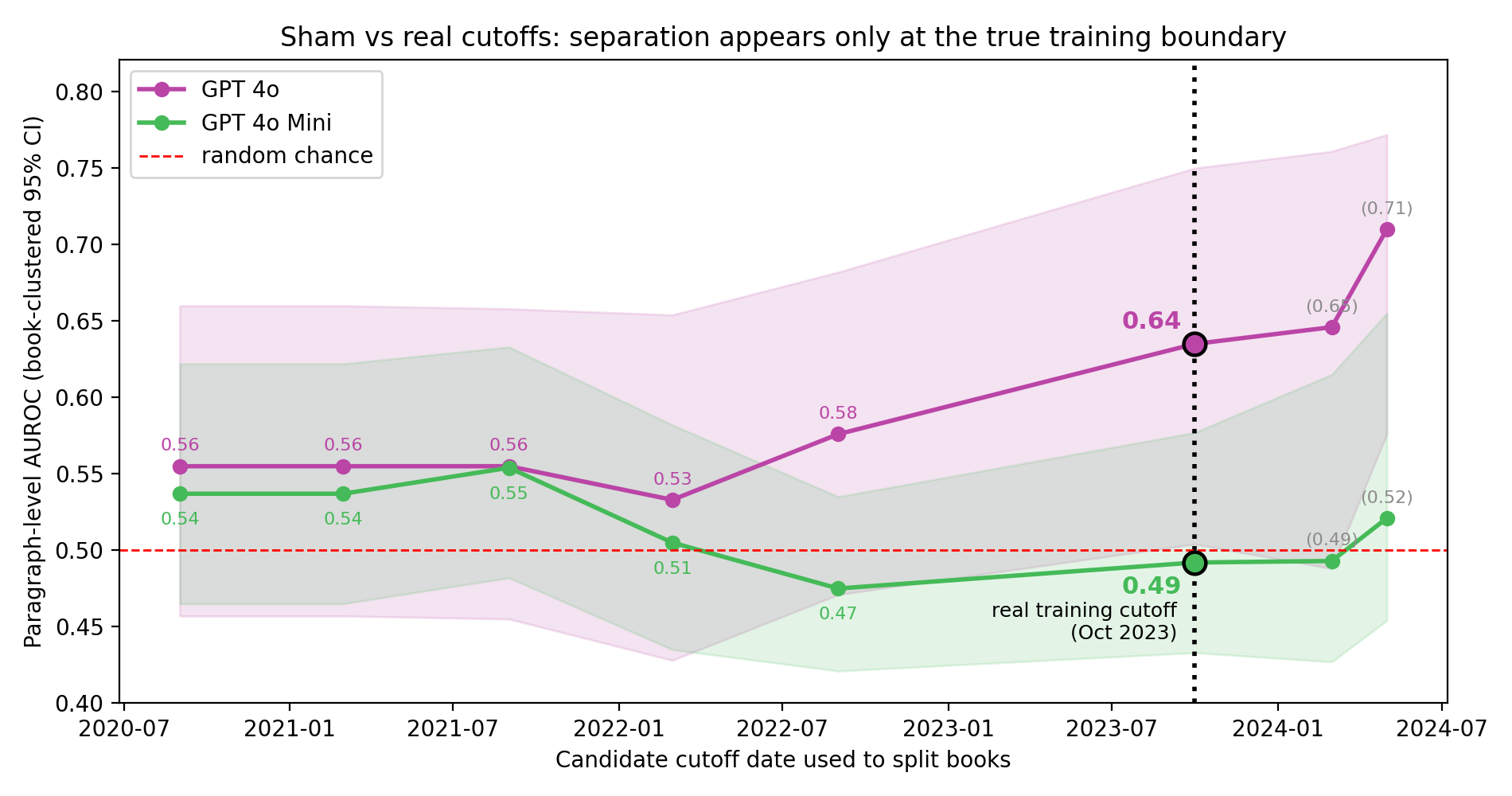}
  \caption*{Note: Paragraph level AUROC (book-clustered bootstrap 95\% CIs) when splitting books at each candidate cutoff date, for GPT-4o and GPT-4o Mini (identical books and paraphrases). Candidates inside the cutoff year are omitted because cutoff-year books are excluded from analysis}
  \label{fig:rolling_sham}
\end{figure}

\textit{The effect of testing fake cutoff dates.} As a temporal-boundary diagnostic, we swept candidate cutoff dates across the main GPT-4o and GPT-4o Mini analysis samples (excluding books published in the cutoff year, as in the main analysis) and computed paragraph level AUROC with book-clustered bootstrap intervals. Candidate cutoffs before October 2023 ask whether other date partitions produce comparable separation. If publication-date style drift were driving the result, we would expect more comparable separation away from the true boundary. It did not: pre-boundary candidate splits yielded paragraph level AUROC scores of 0.53--0.58. The real cutoff (October 2023) produced a higher score (0.635, CI 0.504--0.750; see Figure \ref{fig:rolling_sham}).

When looking at fake cutoff dates after the real cutoff date (cutoff dates that will classify unseen books as seen) we found that some fake cutoffs are elevated above the real one. After further investigation we found that two books that we classified as unseen due to their publication date were partially available before the cutoff through O'Reilly's Early Release program (\ref{sec:partially-seen}).

\textit{Permutation null.} Finally, we compared each model's observed non-public book level AUROC against the null distribution obtained by shuffling book labels 5,000 times. GPT-4o's non-public AUROC of 0.82 lies in the upper tail (one-sided p = 0.014; null 95th percentile 0.75), whereas GPT-4o Mini (p = 0.36) and GPT-3.5 Turbo (p = 0.38) are indistinguishable from label noise (see Table \ref{tab:permutation_null}).

\begin{table}[H]
\centering
\renewcommand{\arraystretch}{1.5}
\caption{\centering {\large{Permutation Null: Observed Book Level AUROC vs.\ 5,000 Label Shuffles}}}
\footnotesize
\begin{adjustbox}{max width=\linewidth}
\begin{tabular}{llcccc}
\toprule
Model & Data-Split & Observed AUROC & One-Sided P-value & Null Mean & Null 95th Pctl. \\
\midrule
\multirow{3}{*}{GPT-4o} & All & 0.79 & 0.026 & 0.50 & 0.75 \\
 & Public & 0.64 & 0.184 & 0.50 & 0.75 \\
 & Non-Public & 0.82 & 0.014 & 0.50 & 0.75 \\
\midrule
\multirow{3}{*}{GPT-4o Mini} & All & 0.56 & 0.361 & 0.50 & 0.75 \\
 & Public & 0.55 & 0.393 & 0.50 & 0.75 \\
 & Non-Public & 0.56 & 0.359 & 0.50 & 0.74 \\
\midrule
\multirow{3}{*}{GPT-3.5 Turbo} & All & 0.58 & 0.274 & 0.50 & 0.71 \\
 & Public & 0.64 & 0.140 & 0.50 & 0.70 \\
 & Non-Public & 0.54 & 0.377 & 0.50 & 0.71 \\
\bottomrule
\end{tabular}
\end{adjustbox}
\normalsize
\caption*{Note: Book labels (pre- vs.\ post-cutoff) are shuffled 5,000 times while book scores are held fixed; the p-value is the fraction of shuffles whose AUROC meets or exceeds the observed value. Only GPT-4o's non-public (and all-paragraph) AUROC lies outside its null distribution.}
\label{tab:permutation_null}
\end{table}

\subsubsection{Leave-one-book-out sensitivity}
\label{sec:lobo}

\begin{figure}[H]
  \centering
  \caption{\centering {\large{Leave-One-Book-Out Sensitivity of Book Level AUROC}}}
  \vspace{-2mm}
  \includegraphics[width=0.95\linewidth]{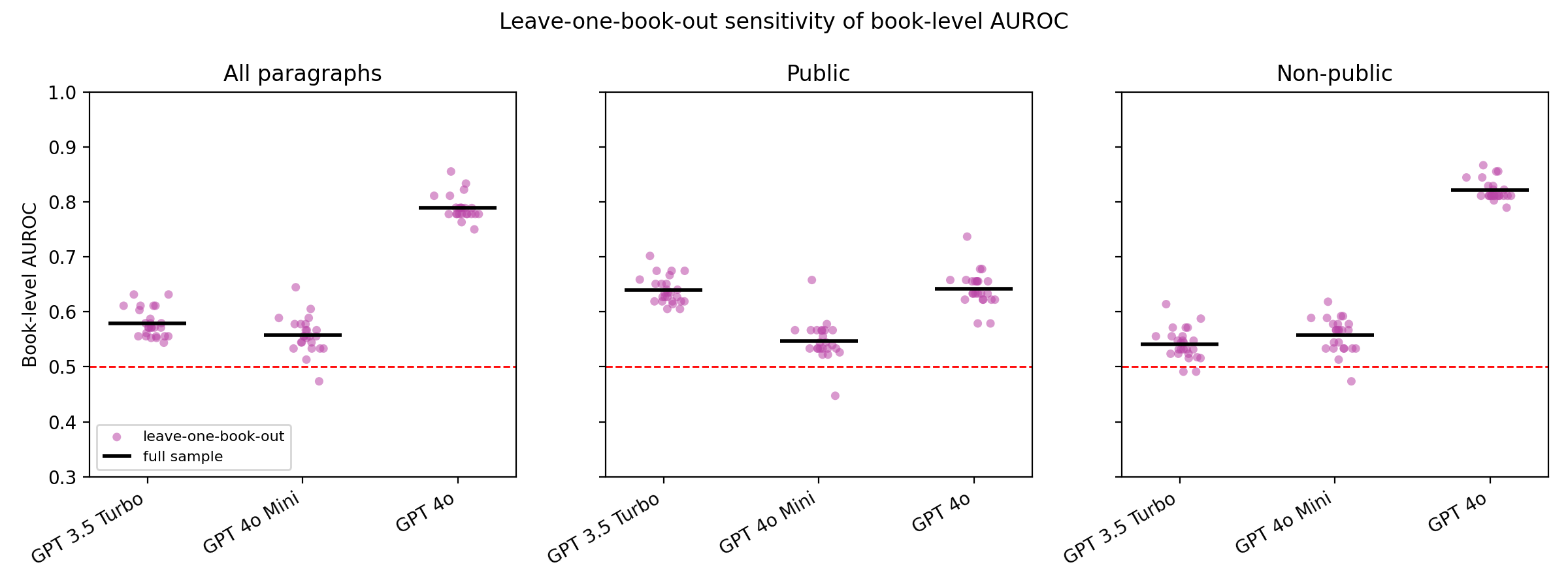}
  \caption*{Note: Each dot is the book level AUROC with one book removed; the black bar is the full-sample value. Per-book values for every model and data split are included in the numerical outputs accompanying our code release.}
  \label{fig:lobo}
\end{figure}

With fewer than 30 books per AUROC estimate, individual titles could have an outsized influence on our results. To check this, we recomputed the book level AUROC with each book excluded in turn. GPT-4o's non-public AUROC ranged from 0.79 to 0.87 around the full sample value of 0.82. Every estimate remained well above chance and above the full sample values of the other models, indicating that the headline result is not driven by any single book (see Figure \ref{fig:lobo}; per-book values are included in the numerical outputs accompanying our code release). The same analysis for the other models and data splits showed similarly modest sensitivity (GPT-4o Mini ranged from 0.47 to 0.65 around 0.56 on all paragraphs; GPT-3.5 Turbo ranged from 0.54 to 0.63 around 0.58). On the all-paragraph split, the exclusion that increases GPT-4o's AUROC the most is \textit{Building Green Software}, one of two formally post-cutoff books that were partially available before the cutoff (see Appendix \ref{sec:partially-seen}).

\subsubsection{Public/private style and position controls}
\label{sec:public-private-controls}

The public/non-public comparison is difficult to interpret causally because the two text splits are structurally different. Public preview text outside chapters one and four consists of roughly the first 1,500 characters of each chapter (about 230 words in the median chapter), while non-public text is mostly body text from the rest of the chapter. The aggregate paired bootstrap above shows a statistically significant public/non-public difference for GPT-4o, but this does not by itself isolate access status from style, position, or chapter structure. This appendix therefore addresses a secondary interpretation of the results: whether the observed public/non-public gap can be read as a clean paywall-status effect. It does not test, or overturn, the paper's primary within-non-public comparison between books published before and after the training cutoff.

We therefore ran two matched local controls. First, for each chapter containing both public and non-public tested snippets, we compared the last public snippet before the preview boundary with the first non-public snippet immediately after it, excluding chapters with no non-public snippets. This boundary-adjacent comparison does not reproduce the aggregate private-text advantage. For GPT-4o, the last-public book level AUROC was 0.80, while the first-private AUROC was 0.74; the paired book-level bootstrap difference was not significant ($-0.06$, 95\% CI: $-0.26$ to 0.11, $p = 0.55$; Table \ref{tab:boundary_adjacent}).

Second, we compared the fully public chapter four with adjacent non-public text from chapters three and five. This is less local than the boundary comparison, but it compares nearby chapters rather than the full public and non-public corpora. For GPT-4o, chapter three/five private text had a higher AUROC than chapter four public text (0.67 vs 0.57), but the paired book-level difference was again not statistically significant (0.11, 95\% CI: $-0.11$ to 0.35, $p = 0.39$; Table \ref{tab:adjacent_chapter}).

Taken together, these matched controls suggest that the full public/non-public gap is an aggregate pattern rather than a clean local effect of the public/private boundary. They do not overturn the main non-public pre-cutoff versus post-cutoff finding, but they do leave stylistic and structural differences between preview and non-preview text as a plausible contributor to the public/non-public contrast. For this reason, we treat the public/non-public comparison as secondary and descriptive, and place greater weight on the within-non-public comparison across the training cutoff.

\begin{table}[H]
\centering
\renewcommand{\arraystretch}{1.5}
\footnotesize
\caption{\centering {\large{Boundary-Adjacent Control Around the Public/Private Preview Cutoff}}}
\begin{tabular}{lrrrrrr}
\toprule
Model & Last Public & First Private & Diff. & 95\% CI & P-value & Pairs \\
\midrule
GPT-4o & 0.80 & 0.74 & $-0.06$ & ($-0.26$, 0.11) & 0.55 & 281 \\
GPT-4o Mini & 0.79 & 0.63 & $-0.16$ & ($-0.44$, 0.11) & 0.28 & 282 \\
GPT-3.5 Turbo & 0.44 & 0.27 & $-0.17$ & ($-0.41$, 0.04) & 0.11 & 225 \\
\bottomrule
\end{tabular}
\normalsize
\caption*{Note: For each chapter with both public and non-public tested snippets, we select the last public snippet before the preview boundary and the first non-public snippet after it. AUROC values are book level. Difference is first-private minus last-public AUROC from a paired book-level bootstrap. Chapters with no non-public snippets are excluded.}
\label{tab:boundary_adjacent}
\end{table}

\begin{table}[H]
\centering
\renewcommand{\arraystretch}{1.5}
\footnotesize
\caption{\centering {\large{Adjacent-Chapter Control: Chapter 4 Public vs Chapters 3 and 5 Private}}}
\begin{tabular}{lrrrrrr}
\toprule
Model & Ch. 4 Public & Ch. 3/5 Private & Diff. & 95\% CI & P-value & Books \\
\midrule
GPT-4o & 0.57 & 0.67 & 0.11 & ($-0.11$, 0.35) & 0.39 & 24 \\
GPT-4o Mini & 0.44 & 0.43 & $-0.01$ & ($-0.20$, 0.19) & 0.97 & 24 \\
GPT-3.5 Turbo & 0.46 & 0.59 & 0.13 & ($-0.17$, 0.45) & 0.42 & 25 \\
\bottomrule
\end{tabular}
\normalsize
\caption*{Note: AUROC values are book level. Difference is chapter three/five private minus chapter four public AUROC from a paired book-level bootstrap. Chapter four is fully public in the O'Reilly preview split; chapters three and five are adjacent non-public chapters when available.}
\label{tab:adjacent_chapter}
\end{table}

\subsubsection{Partially seen books}
\label{sec:partially-seen}

Prompted by the elevated post-boundary values in the rolling sham analysis (Appendix \ref{sec:placebo}), we audited the availability of the five books published after GPT-4o's training cutoff. Two of them were partially available before the cutoff through O'Reilly's Early Release program, which publishes draft chapters on the O'Reilly platform ahead of formal publication: \textit{Building Green Software} (formally published March 2024) was available from July 2023, and \textit{Deciphering Data Architectures} (formally published February 2024) was available from May 2023, with nine of its sixteen chapters online by September 2023.\footnote{The early release listing for \textit{Building Green Software} is dated July 6, 2023. The author of \textit{Deciphering Data Architectures} documented its early release timeline on his blog: the first two chapters were available by May 2023 and nine chapters by September 2023. The remaining chapters appeared during October and November 2023, around and after the cutoff, so we count only the September milestone as clearly pre-cutoff.} Their content may therefore have been available for training despite their 2024 publication dates. We found no pre-cutoff early release availability for the other three post-cutoff books.

Consistent with this, \textit{Building Green Software}'s identification rate is as high as that of the suspected member books, and dropping it produces the largest single-book increase on the all-paragraph split in the leave-one-book-out analysis (Appendix \ref{sec:lobo}). If these two books were trained on, treating them as unseen makes our estimates conservative, since a trained-on book sitting in the unseen class lowers the measured AUROC: excluding \textit{Building Green Software} alone raises GPT-4o's book level AUROC from 0.79 to 0.86.

\subsection{Prompts and Settings Used for Tested Models}
\label{sec:prompts_and_settings_openai}

To test the model's ability to differentiate between member and non-member data, we used a modified version of the prompts presented in \cite{decop2024}. An example prompt is shown below.

\bigskip
\noindent\textbf{System Prompt:}
\begin{Verbatim}
You are a helpful assistant. You must answer using only the
provided options A, B, C, or D, you may not decline to answer.
\end{Verbatim}

\bigskip
\noindent\textbf{User Prompt:}
\begin{Verbatim}
Question: Which of the following passages is verbatim from
the "97 Things Every Information Security Professional
Should Know" book by Christina Morillo?

Options:
A. Learn with a community. I've personally found that I've
developed new and existing skills much faster (by a
magnitude of months) when I've joined a community of
learners than when I was trying to learn the same skills
alone. By attending local and online user groups,
conferences, and other events, you can discover new
concepts, hone new skills, and network with possible
future colleagues. Also, in a community, you will gain a
more holistic perspective of information security and a
more complete picture of how others are managing
successful information security programs.

B. Connect with fellow learners. From my experience,
acquiring and improving abilities has been significantly
quicker (saving several months) when participating in group
learning compared to solo studying. Going to regional and
virtual meetups, seminars, and similar gatherings helps
you explore fresh ideas, develop capabilities, and build
relationships with potential workmates. Furthermore,
learning within a group provides broader insights into
cybersecurity and better understanding of how various
organizations implement effective security initiatives.

C. Join a learning group. Based on my observations,
mastering both new and current abilities happens much more
rapidly (reducing learning time by months) when I'm part
of a learning circle versus studying independently. Through
participation in area-based and internet-hosted gatherings,
symposiums, and other meetings, you'll encounter different
concepts, sharpen your abilities, and connect with
prospective professional contacts. Additionally, group
involvement offers deeper understanding of security
practices and clearer insights into successful security
program management across organizations.

D. Engage in collaborative learning. My personal journey
shows that skill acquisition and enhancement occurs
substantially faster (cutting months off learning time)
within group settings rather than individual efforts. By
taking part in both physical and digital group meetings,
industry events, and related activities, you can learn new
approaches, improve your capabilities, and establish
connections with future professional peers. Moreover,
group settings provide comprehensive knowledge about
information security and valuable examples of how different
teams run successful security operations.

Answer:
\end{Verbatim}

Our prompt differs slightly from the prompt used in the DE-COP paper. We changed it to help ensure that the model follows the instructions and returns the tokens needed in the limited token log probabilities that OpenAI provides.

\bigskip
\noindent\textbf{Model settings}:
\begin{Verbatim}
{
    "max_tokens": 1,
    "temperature": 0,
    "seed": 2319,
    "logprobs": True,
    "logit_bias": {32: +100, 33: +100, 34: +100, 35: +100},
    "top_logprobs": 20
}
\end{Verbatim}

The exact models tested were as follows: gpt-4o-2024-08-06, gpt-4o-mini-2024-07-18 and gpt-3.5-turbo-1106.
\subsection{Prompts and Settings Used for Paraphrase model}
\label{sec:prompts_and_settings_claude}

We used Claude 3.5 Sonnet (and two other models for robustness) to generate paraphrases from the O'Reilly Media books. An example prompt is shown below.

\bigskip
\noindent\textbf{User Prompt:}
\begin{Verbatim}
Rewrite this entire text (all sentences with no exception)
expressing the same meaning using different words. Aim to keep the
rewriting similar in length to the original text. Do it three
times. The text to be rewritten is identified as <Example A>.
Format your output as:
Example B: <insert paraphrase B>
Example C: <insert paraphrase C>
Example D: <insert paraphrase D>
-
Example A: In general, a soft trade-off exists between active learning that's
useful for maximally improving your model globally and active
learning that's useful for maximizing the likelihood that a user can
and will rate a particular item. Let's look at one particular example
that uses both.
\end{Verbatim}

\bigskip
\noindent\textbf{Model settings}:
\begin{Verbatim}
{
    "temperature": 0.1,
    "model":"claude-3.5-sonnet"
}
\end{Verbatim}

\end{document}